\documentclass{article} % For LaTeX2e
\usepackage{nips15submit_e,times}
\usepackage{hyperref}
\usepackage{url}

\usepackage{amsmath}
\usepackage{amsthm}
\usepackage{amssymb}
\usepackage{graphicx}
\usepackage{xspace}
\usepackage{tabularx}
\usepackage{multicol}
\usepackage{multirow}
\usepackage{url}
\usepackage{wrapfig}
\usepackage{comment}
\usepackage{listings}
\usepackage[utf8]{inputenc}
\usepackage{fancyvrb}
\usepackage{booktabs}
\usepackage{color,soul}
\usepackage[normalem]{ulem}

\newcommand{\bmx}[0]{\begin{bmatrix}}
\newcommand{\emx}[0]{\end{bmatrix}}

\newcommand{\vect}[1]{\mathbf{#1}}
\newcommand{\vects}[1]{\boldsymbol{#1}}
\newcommand{\matr}[1]{\mathbf{#1}}

\newcommand{\vh}[0]{\vect{h}}

\newcommand{\vx}[0]{\vect{x}}

\newcommand{\mI}{\matr{I}}

\newcommand{\TT}[0]{\vects{\theta}}

\newcommand{\vzero}[0]{\vect{0}}

\newcommand{\NN}[0]{\mathcal{N}}
\newcommand{\LL}[0]{\mathcal{L}}

\newcommand{\RR}[0]{\mathbb{R}}

\newcommand{\E}[0]{\matr{E}}

\DeclareMathOperator*{\argmax}{\arg \max}
\DeclareMathOperator*{\argmin}{\arg \min}

\title{Noisy Parallel Approximate Decoding \\ for Conditional Recurrent Language
Model}

\author{
    Kyunghyun Cho \\
New York University \\
\texttt{kyunghyun.cho@nyu.edu} 
}

\nipsfinalcopy % Uncomment for camera-ready version

\begin{document}

\maketitle

\begin{abstract}
    Recent advances in conditional recurrent language modelling have mainly
    focused on network architectures (e.g., attention mechanism), learning
    algorithms (e.g., scheduled sampling and sequence-level training) and novel
    applications (e.g., image/video description generation, speech recognition,
    etc.) On the other hand, we notice that decoding algorithms/strategies have
    not been investigated as much, and it has become standard to use greedy or
    beam search. In this paper, we propose a novel decoding strategy motivated
    by an earlier observation that nonlinear hidden layers of a deep neural
    network stretch the data manifold. The proposed strategy is embarrassingly
    parallelizable without any communication overhead, while improving an
    existing decoding algorithm. We extensively evaluate it with attention-based
    neural machine translation on the task of En$\to$Cz translation.
\end{abstract}

\section{Introduction}

Since its first use as a language model in 2010 \cite{mikolov2010recurrent}, a
recurrent neural network has become a {\it de facto} choice for implementing a
language model~\cite{zoph2016rnn,sundermeyer2015feedforward}.  One of the
appealing properties of this approach to language modelling, to which we refer
as {\it recurrent language modelling}, is that a recurrent language model can
generate a long, coherent sentence~\cite{sutskever2011generating}. This is due
to the ability of a recurrent neural network to capture long-term dependencies.

This property has come under spotlight in recent years as the conditional
version of a recurrent language model began to be used in many different
problems that require generating a natural language description of a
high-dimensional, complex input. These tasks include machine translation, speech
recognition, image/video description generation and many more
\cite{cho2015describing} and references therein.

Much of the recent advances in conditional recurrent language model have focused
either on network architectures (e.g.,
\cite{bahdanau2014neural}), learning algorithms (e.g.,
\cite{bengio2015scheduled,ranzato2015sequence,bahdanau2015task}) or novel
applications (see \cite{cho2015describing} and references therein). On the other
hand, we notice that there has not been much research on decoding algorithms for
conditional recurrent language models. In the most of work using recurrent
language models, it is a common practice to use either greedy or beam search to
find the most likely natural language description given an input.

In this paper, we investigate whether it is possible to decode better from a
conditional recurrent language model. More specifically, we propose a decoding
strategy motivated by earlier observations that nonlinear hidden layers of a
deep neural network stretch the data manifold such that a neighbourhood in the
hidden state space corresponds to a set of semantically similar configurations
in the input space \cite{bengio2013better}. This observation is exploited in the
proposed strategy by injecting noise in the hidden transition function of a
recurrent language model. 

The proposed strategy, called noisy parallel approximate decoding (NPAD), is a
meta-algorithm that runs in parallel many chains of the noisy version of an
inner decoding algorithm, such as greedy or beam search. Once those parallel
chains generate the candidates, the NPAD selects the one with the highest score.
As there is effectively no communication overhead during decoding, the
wall-clock performance of the proposed NPAD is comparable to a single run of an
inner decoding algorithm in a distributed setting, while it improves the
performance of the inner decoding algorithm. We empirically evaluate the
proposed NPAD against the greedy search, beam search as well as stochastic
sampling and diverse decoding \cite{li2016mutual} in attention-based
neural machine translation.

\section{Conditional Recurrent Language Model}

A language model aims at modelling a probabilistic distribution over natural
language text. A recurrent language model is a language model implemented as a
recurrent neural network \cite{Mikolov-thesis-2012}.

Let us define a probability of a given natural language sentence,\footnote{
    Although I use a ``sentence'' here, this is absolutely not necessary, and
    any level of text, such as a phrase, paragraph, chapter and document, can be
    used as a unit of language modelling. Furthermore, it does not have to be a
    natural language text but any sequence such as speech, video or actions.
}
which we represent
as a sequence of linguistic symbols $X=(x_1, x_2, \ldots, x_T)$, as
\begin{align}
    \label{eq:p_x}
    p(X) = p(x_1, x_2, \ldots, x_T) = p(x_1) p(x_2|x_1) p(x_3|x_1,x_2) \cdots
    p(x_T | x_{<T}) = \prod_{t=1}^T p(x_t|x_{<t}),
\end{align}
where $x_{<t}$ is all the symbols preceding the $t$-th symbol in the sentence $X$.
Note that this conditional dependency structure is not necessary but is
preferred over other possible structures due to its naturalness as well as the
fact that the length of a given sentence $T$ is often unknown in advance.

In a neural language model~\cite{bengio2003neural}, a neural network is used to
compute each of the conditional probability terms in Eq.~\eqref{eq:p_x}. A
difficulty in doing so is that the input $(x_1, x_2, \ldots, x_{t-1})$ to the
neural network is of variable size. A recurrent neural network cleverly
addresses this difficulty by reading one symbol at a time while maintaining an
internal memory state:
\begin{align}
    \label{eq:rnn}
    \vh_t = \phi\left( \vh_{t-1}, \E\left[x_t\right] \right),
\end{align}
where $\vh_t$ is the internal memory state at time $t$. $\E\left[x_t\right]$ is a vector
representation of the $t$-th symbol in the input sentence. The internal memory state
$\vh_t$ effectively summarizes all the symbols read up to the $t$-th time step.

The recurrent activation function $\phi$ in Eq.~\eqref{eq:rnn} can be as simple
as an affine transformation followed by a point-wise nonlinearity (e.g.,
$\tanh$) to as complicated a function as long short-term memory (LSTM,
\cite{hochreiter1997long}) or gated recurrent units (GRU,
\cite{cho2014learning}). The latter two are often preferred, as they effectively
avoid the issue of vanishing gradient~\cite{bengio1994learning}.

Given the internal hidden state, the recurrent neural network computes the
conditional distribution over the next symbol $x_{t+1}$. Assuming a fixed
vocabulary $V$ of linguistic symbols, it is straightforward to make a parametric
function that returns a probability of each symbol in the vocabulary:
\begin{align}
    \label{eq:output}
    p(x_{t+1}=j | x_{\leq t}) = 
    \frac{\exp(g_j(\vh_t))}{\sum_{j'=1}^{|V|} \exp(g_{j'}(\vh_t))},
\end{align}
where $g_j(\vh_t)$ is the $j$-th component of the output of the function
$g:\RR^{\dim(\vh_t)}\to\RR^{|V|}$. The formulation on the right-hand side of
Eq.~\eqref{eq:output} is called a softmax function~\cite{bridle1990probabilistic}.

Given Eqs.~\eqref{eq:rnn}--\eqref{eq:output}, the recurrent neural network reads
one symbol of a given sentence $X$ at a time from left to right and computes the
conditional probability of each symbol until the end of the sequence is reached.
The probability of the sentence is then given by a product of all those conditional
probabilities. We call this recurrent neural network a {\it recurrent language
model}.

\paragraph{Conditional Recurrent Language Model}

A recurrent language model is turned into a {\it conditional recurrent language
model}, when the distribution over sentences is conditioned on another modality
including another language. In other words, a conditional recurrent language
model estimates
\begin{align}
    \label{eq:crlm}
    p(X|Y) = \prod_{t=1}^T p(x_t | x_{<t}, Y).
\end{align}

$Y$ in Eq.~\eqref{eq:crlm} can be anything from a sentence in another language
(machine translation), an image (image caption generation), a video clip (video
description generation) to speech (speech recognition). In any of those cases, a
previously described recurrent language model requires only a slightest tweak in
order to take into account $Y$. 

The tweak is to compute the internal hidden state of the recurrent language
model based not only on $\vh_{t-1}$ and $\E\left[x_t\right]$ (see Eq.~\eqref{eq:rnn}) but
also on $Y$ such that
\begin{align}
    \label{eq:rnn_cond}
    \vh_t = \phi\left( \vh_{t-1}, \E\left[x_t\right], f(Y, t) \right),
\end{align}
where $f$ is a time-dependent function that maps from $Y$ to a vector.
Furthermore, we can make $g_j$ in Eq.~\eqref{eq:output} to be conditioned on $Y$
as well
\begin{align}
    \label{eq:output_cond}
    p(x_{t+1}=j | x_{\leq t}) = 
    \frac{\exp(g_j(\vh_t, f(Y, t)))}{\sum_{j'=1}^{|V|} \exp(g_{j'}(\vh_t, f(Y, t)))}.
\end{align}

\paragraph{Learning}
Given a data set $D$ of pairs $(X, Y)$, the conditional recurrent language model
is trained to maximize the log-likelihood function which is defined as
\begin{align*}
    \LL(\TT) = \frac{1}{|D|} \sum_{n=1}^N \sum_{t=1}^{T^n} \log
    p(x_t^n|x_{<t}^n, Y^n).
\end{align*}
This maximization is often done by stochastic gradient descent with the gradient
computed by backpropagation~\cite{rumelhart1986learning}. Instead of a scalar
learning rate, adaptive learning rate methods, such as
Adadelta~\cite{zeiler2012adadelta} and
Adam~\cite{kingma2014adam}, are often used.

\section{Decoding}
\label{sec:decoding}

Decoding in a conditional recurrent language model corresponds to finding a
target sequence $\tilde{X}$ that maximizes the conditional probability $p(X|Y)$
from Eq.~\eqref{eq:crlm}:
\begin{align*}
    \tilde{X} = \argmax_X \log p(X | Y).
\end{align*}
As is clear from the formulation in
Eqs.~\eqref{eq:rnn_cond}--\eqref{eq:output_cond}, exact decoding is intractable,
as the state space of $X$ grows exponentially with respect to the length of the
sequence, i.e., $|\mathcal{X}|=O(|V|^{|X|})$, without any trivial structure that
can be exploited. Thus, we must resort to approximate decoding. 

\subsection{Greedy Decoding}

Greedy decoding is perhaps the most naive way to approximately decode from the
conditional recurrent language model. At each time step, it greedily selects the
most likely symbol under the conditional probability:
\begin{align}
    \label{eq:greedy}
    \tilde{x}_t = \argmax_j \log p(x_t = j | \tilde{x}_{<t}).
\end{align}
This continues until a special marker indicating the end of the sequence is
selected. 

This greedy approach is computationally efficient, but is likely too crude. Any
early choice based on a high conditional probability can easily turn out to be
unlikely one due to low conditional probabilities later on. This issue is
closely related to the garden path sentence problem~(see Sec.~3.2.4 of
\cite{manning1999foundations}.)

%\paragraph{Computational Complexity}
%
%When the maximum length of $X$ is set to $T$, the computational complexity of
%the greedy decoding is $O((|V|+C)T)$. $|V|$ comes from the fact that each update
%\eqref{eq:greedy} requires the algorithm to find the max of $|V|$
%log-probabilities. $C$ is the cost of computing
%Eqs.~\eqref{eq:rnn_cond}--\eqref{eq:output_cond}.

\subsection{Beam Search}

Beam search improves upon the greedy decoding strategy by maintaining $K$
hypotheses at each time step, instead of a single one. Let 
\[
    \mathcal{H}_{t-1} = \left\{ 
        (\tilde{x}_1^1, \tilde{x}_2^1, \ldots, \tilde{x}_{t-1}^1), 
        (\tilde{x}_1^2, \tilde{x}_2^2, \ldots, \tilde{x}_{t-1}^2), 
        \ldots, 
        (\tilde{x}_1^K, \tilde{x}_2^K, \ldots, \tilde{x}_{t-1}^K)
    \right\}
\]
be a set of current hypotheses at time $t$. Then, from each current hypothesis
the following $|V|$ candidate hypotheses are generated:
\begin{align*}
    \mathcal{H}^k_t = \left\{
        (\tilde{x}_1^k, \tilde{x}_2^k, \ldots, \tilde{x}_{t-1}^k, v_1), 
        (\tilde{x}_1^k, \tilde{x}_2^k, \ldots, \tilde{x}_{t-1}^k, v_2), 
        \ldots, 
        (\tilde{x}_1^k, \tilde{x}_2^k, \ldots, \tilde{x}_{t-1}^k, v_{|V|})
    \right\},
\end{align*}
where $v_j$ denotes the $j$-th symbols in the vocabulary $V$.

The top-$K$ hypotheses from the union of all such hypotheses sets
$\mathcal{H}^k_t,k=1,\ldots,K$ are selected based on their scores. In other
words,
\begin{align*}
    \mathcal{H}_t = \cup_{k=1}^K \mathcal{B}_k,
\end{align*}
where 
\begin{align*}
    \mathcal{B}_k = \argmax_{\tilde{X} \in
    \mathcal{A}_{k}} \log p(\tilde{X} | Y),~
    \mathcal{A}_k = \mathcal{A}_{k-1} - \mathcal{B}_{k-1},\text{ and }
    \mathcal{A}_1 = \cup_{k'=1}^K \mathcal{H}^{k'}_t.
\end{align*}

Among the top-$K$ hypotheses, we consider the ones whose last symbols are the
special marker for the end of sequence to be complete and stop expanding such
hypotheses. All the other hypotheses continue to be expanded, however, with $K$
reduced by the number of complete hypotheses. When $K$ reaches $0$, the beam
search ends, and the best one among all the complete hypotheses 
is returned.

%\paragraph{Computational Complexity}
%
%The computational complexity of the beam search strategy is $O((|V|K\log K +
%KC)T)$. $T$ is as before the maximum length of $X$, and $|V|$ is the size of the
%symbol dictionary. The term $|V|K \log K$ comes from selecting the top-$K$
%hypotheses among $|V|\times K$ candidate hypotheses, and $KC$ from running a
%single step of the recurrent neural network for $K$ hypotheses.

\section{NPAD: Noisy Parallel Approximate Decoding}

In this section, we introduce a strategy that can be used in conjunction
with the two decoding strategies discussed earlier. This new strategy is
motivated by the fact that a deep neural network, including a recurrent neural
network, learns to stretch the input manifold (on which only likely input
examples lie) and fill the hidden state space with it. This implies that a
neighbourhood in the hidden state space corresponds to a set of semantically
similar configurations in the input space, regardless of whether those
configurations are close to each other in the input space
\cite{bengio2013better}.  In other words, small perturbation in
the hidden space corresponds to jumping from one plausible
configuration to another. 

In the case of conditional recurrent language model, we can achieve this
behaviour of efficiently exploration across multiple modes by injecting noise to
the transition function of the recurrent neural network. In other words, we
replace Eq.~\eqref{eq:rnn_cond} with
\begin{align}
    \label{eq:npad_noise}
    \vh_t = \phi\left( \vh_{t-1} + \epsilon_{t}, \E\left[ x_t\right], f(Y, t) \right),
\end{align}
where 
\[
    \epsilon_t \sim \NN(\vzero, \sigma_t^2 \mI).
\]

The time-dependent standard deviation $\sigma_t$ should be selected to reflect
the uncertainty dynamics in the conditional recurrent language model. As the
recurrent network models a target sequence in one direction, uncertainty is
often greatest when predicting earlier symbols and gradually decreases as more
and more context becomes available for the conditional distribution $p(y_t |
y_{<t})$. This naturally suggests a strategy where we start with a high level of
noise (high $\sigma_t$) and anneal it ($\sigma_t \to 0$) as the decoding
progresses. One such scheduling scheme is 
\[
    \sigma_t = \frac{\sigma_0}{t},
\]
where $\sigma_0$ is an initial noise level. Although there are many
alternatives, we find this simple formulation to be effective in experiments
later.

We run $M$ such noisy decoding processes in parallel. This can be done easily
and efficiently, as there is no communication between these parallel processes
except at the end of the decoding processing. Let us denote by $\tilde{Y}_m$ a
sequence decoded from the $m$-th decoding process. Among these $M$ hypotheses,
we select the one with the highest probability assigned by the {\it
non-noisy} model:
\[
    \tilde{Y} = \argmax_{\tilde{Y}_m: m=1,\ldots,M} \log p(\tilde{Y}_m | X).
\]

We call this decoding strategy, based on running multiple parallel approximate
decoding processes with noise injected, {\it noisy parallel approximate
decoding} (NPAD). 

\paragraph{Computational Complexity}
Clearly, the proposed decoding strategy is $M$ times more expensive, i.e.,
$O(MD)$, where $D$ is the computational complexity of either greedy or beam
search (see Sec.~\ref{sec:decoding}.) It is however important to note that the
proposed NPAD is embarrassingly parallelizable, which is well suited for
distributed and parallel environments of modern computing. By utilizing
multi-core machines, the practical cost of computation reduces to simply running
the greedy or beam search once (with a constant multiplicative factor of
$2\pm\epsilon$ due to computing the non-noisy score and generating pseudo random
numbers.) This is contrary to, for instance, when comparing the beam search to
the greedy search, in which case the benefit from parallelization is limited due
to the heavy communication cost at each step.

\paragraph{Quality Guarantee}
A major issue with the proposed strategy is that the resulting sequence may be worse
than running a single inner-decoder, due to the stochasticity. This is however
easily avoided by setting $\sigma_0$ to $0$ for one of the $M$ decoding
processes. By doing so, even if all the other noisy decoding processes resulted
in sequences whose probabilities are worse than the non-noisy process, the
proposed strategy nevertheless returns a sequence that is as good as a single run
of the inner decoding algorithm.

\subsection{Why not Sampling?}

The formulation of the conditional recurrent language model in
Eq.~\eqref{eq:crlm} implies that we can generate exact samples from the model,
as this is a directed acyclic graphical model. At each time step $t$, a sample
from the categorical distribution given all the samples of the previous time
steps (Eq.~\eqref{eq:output_cond}) is generated. This procedure is done
iteratively either up to $T$ time steps or another type of stopping criterion is
met (e.g., the end-of-sequence symbol is sampled.) Similarly to the proposed
NPAD, we can run a set of this sampling procedures in parallel. 

A major difference between this sampling-at-the-output and the proposed NPAD is
that the NPAD exploits the hidden state space of a neural network
in which the data manifold is highly {\it linearized}. In other words, training
a neural network tends to {\it fill up} the hidden state space as much as
possible with valid data points,\footnote{
    This behaviour can be further encouraged by regularizing the
    (approximate) posterior over the hidden state, for instance, as in
    variational autoencoders~(see, e.g., \cite{kingma2013auto,chung2015recurrent}.)
}
and consequently any point in the neighbourhood of a valid
hidden state ($\vh_t$ Eq.~\eqref{eq:rnn_cond}) should map to a plausible point
in the output space. This is contrary to the actual output space, where only a
fraction of the output space is plausible.

Later, we show empirically that it is indeed more efficient to sample in the
hidden state space than in the output state space.

\subsection{Related Work}

\paragraph{Perturb-and-MAP}

Perturb-and-MAP~\cite{papandreou2011perturb} is an algorithm that reduces
probabilistic inference, such as sampling, to energy minimization in a Markov
random field (MRF) \cite{papandreou20147}. For instance, instead of Gibbs
sampling, one can use the perturb-and-MAP algorithm to find multiple instances
of configurations that minimize the {\it perturbed} energy function. Each
instance of the perturb-and-MAP works by first injecting noise to the energy
function of the MRF, i.e., $\tilde{E}(\vx) = E(\vx) + \epsilon(\vx)$, followed
by maximum-a-posterior (MAP) step, i.e., $\argmin_{\vx} \tilde{E}(\vx)$. 

A connection between this perturb-and-MAP and the proposed NPAD is clear.  Let
us define the energy function of the conditional recurrent language model as its
log-probability, i.e., $E(X|Y) = \log p(X|Y)$ (see Eq.~\eqref{eq:crlm}.) Then,
the noise injection to the hidden state in Eq.~\eqref{eq:npad_noise} is a
process similar to injecting noise to the energy function. This connection
arises from the fact that the NPAD and perturb-and-MAP share the same goal of
``[giving] other low energy states the chance''~\cite{papandreou20147}.

\paragraph{Diverse Decoding}

One can view the proposed NPAD as a way to generate a diverse set of likely
solutions from a conditional recurrent language model. In \cite{li2016mutual}, a
variant of beam search was proposed, which modifies the scoring function at each
time step of beam search to promote diverse decoding. This is done by penalizing
low ranked hypotheses that share a previous hypothesis. This approach is
however only applicable to beam search and is not as parallelizable as the
proposed NPAD. It should be noted that the NPAD and the diverse decoding can be
used together.

Earlier, Batra~et~al.~\cite{batra2012diverse} proposed another approach that
enables decoding multiple, diverse solutions from an MRF. This method decodes
one solution at a time, while regularizing the energy function of an MRF with
the diversity measure between the solution currently being decoded and all the
previous solutions. Unlike the perturb-and-MAP or the NPAD, this is a
deterministic algorithm. A major downside to this approach is that it is
inherently sequential. This makes it impractical especially for neural machine
translation, as already the major issue behind its deployment is the
computational bottleneck in decoding.

\section{Experiments: Attention-based Neural Machine Translation}

\subsection{Settings}

In this paper, we evaluate the proposed noisy parallel approximate decoding
(NPAD) strategy in attention-based neural machine translation. More
specifically, we train an attention-based encoder-decoder network on the task of
English-to-Czech translation and evaluate different decoding strategies.

The encoder is a single-layer bidirectional recurrent neural network with
1028 gated recurrent units (GRU, \cite{cho2014learning}).\footnote{
    The number 1028 resulted from a typo, when originally we intended to use
    1024.
} The decoder consists of an attention mechanism~\cite{bahdanau2014neural} and a
recurrent neural network again with 1028 GRU's. Both source and target words
were projected to a 512-dimensional continuous space. We used the code from
dl4mt-tutorial available online\footnote{
    \url{https://github.com/nyu-dl/dl4mt-tutorial/tree/master/session2}
} for training. Both source and target sentences were represented as sequences
of BPE subword symbols~\cite{sennrich2015neural}.

We trained this model on a large parallel corpus of approximately 12m
sentence pairs, available from WMT'15,\footnote{
    http://www.statmt.org/wmt15/translation-task.html
}
for 2.5 weeks.  During training,
ADADELTA~\cite{zeiler2012adadelta} was used to adaptively adjust the learning
rate of each parameter, and the norm of the gradient was renormalized to $1$, if
it exceed $1$. The training run was early-stopped based on the validation
perplexity using newstest-2013 from WMT'15. The model is tested with two held-out sets, newstest-2014 and
newstest-2015.\footnote{
    Due to the space constraint, we only report the result on newstest-2014. We
    however observed the same trend from newstest-2014 on newstest-2015.
}

We closely followed the training and test strategies from \cite{firat2016multi},
and more details can be found in it.

\paragraph{Evaluation Metric}

The main evaluation metric is the negative conditional log-probability of a
decoded sentence, where lower is better. Additionally, we use BLEU as a
secondary evaluation metric. BLEU is a de-facto standard metric for
automatically measuring the translation quality of machine translation systems,
in which case higher is better.

\subsection{Decoding Strategies}

We evaluate four decoding strategies. We choose the strategies that have
comparable computational complexity per core/machine, assuming multiple
cores/machines are available. This selection left us with greedy search, beam
search, stochastic sampling, diverse decoding and the proposed NPAD.

\paragraph{Greedy and Beam Search} Both greedy and beam search are the most
widely used decoding strategies in neural machine translation, as well as
other conditional recurrent language models for other tasks. In the case of beam search, we test
with two beamwidths, 5 and 10. We use the script made available at
dl4mt-tutorial.

\paragraph{Stochastic Sampling}
A naive baseline for injecting noise during decoding is to simply sample from
the output distribution at each time step, instead of taking the top-$K$
entries. We test three configurations, where 5, 10 or 50 such samplers are run
in parallel. 

\paragraph{Noisy Parallel Approximate Decoding (NPAD)}

We extensively evaluate the NPAD by varying the number of parallel decoding (5,
10 or 50), the beamwidth (1, 5 or 10) and the initial noise level $\sigma_0$
($0.1$, $0.2$, $0.3$ or $0.5$).

\paragraph{Diverse Decoding}
We try the diverse decoding strategy from \cite{li2016mutual}. There is one
hyperparameter $\eta$, and we search over $\left\{ 0.001, 0.01, 0.1, 1\right\}$,
as suggested by the authors of \cite{li2016mutual} based on the validation set
performance.\footnote{
    Personal communication.
}
Also, we vary the beam width (5 or 10). This is included as a
deterministic counter-part to the NPAD.

\begin{table}[h]
    \small 
    \centering
    \begin{minipage}{0.65\textwidth}
\begin{tabular}{c c || c c | c c }%| c c}
        &             & \multicolumn{2}{c|}{Valid} & \multicolumn{2}{c}{Test-1} \\ %& \multicolumn{2}{c}{Test-2} \\ 
        Strategy & $\sigma_0$ & NLL$\downarrow$ & BLEU$\uparrow$ &
        NLL$\downarrow$ & BLEU$\uparrow$ \\ %& NLL$\downarrow$ & BLEU$\uparrow$ \\
        \hline\hline
        Greedy & - & 27.879 & 15.5 & 26.4928 & 16.66 \\ %& 24.0416 & 13.68 \\
        \hline
        Sto. Sampling & - & 22.9818 & 15.64 & 26.2536 & 16.76 \\ %& 23.8157 & 13.84 \\
        \hline
        NPAD & 0.1 &   21.125  &  16.06  &   23.8542 &  17.48  \\ %&   21.6545 & 14.71 \\
        NPAD & 0.2 &   20.6353 &  16.37  &   23.2631 &  17.86  \\ %&   21.0311 & 14.97 \\
        NPAD & \underline{0.3} &   {\bf 20.4463} &  {\bf 16.71}  & \underline{23.0111} &  \underline{18.03} \\ %&   \underline{20.8868} & \underline{14.98} \\
        NPAD & 0.5 &   20.7648 &  16.48  &   23.3056 &  18.13 %& 21.1847 & 14.88 
    \end{tabular}
\end{minipage}
\begin{minipage}{0.34\textwidth}
    \caption{Effect of noise injection. For both stochastic sampling and NPAD,
    we used 50 parallel samplers. For NPAD, we used the greedy decoding as an
inner-decoding strategy. }
    \label{tab:noise_injection}
\end{minipage}
\end{table}

\subsection{Results and Analysis}

\paragraph{Effect of Noise Injection}

First, we analyze the effect of noise injection by comparing the stochastic
sampling and the proposed NPAD against the deterministic greedy decoding. In
doing so, we used 50 parallel decoding processes for both stochastic sampling
and NPAD. We varied the amount of initial noise $\sigma_0$ as well.

In Table~\ref{tab:noise_injection}, we present both the average negative
log-probability and BLEU for all the cases. As expected, the proposed NPAD
improves upon the deterministic greedy decoding as well as the stochastic
sampling strategy. It is important to notice that the improvement by the NPAD
is significantly larger than that by the stochastic sampling, which confirms
that it is more efficient and effective to inject noise in the hidden state of
the neural network. 

\begin{table}[h]
    \small 
    \centering
    \begin{minipage}{0.65\textwidth}
    \begin{tabular}{c c || c c | c c}% | c c}
        &             & \multicolumn{2}{c|}{Valid} & \multicolumn{2}{c}{Test-1} \\ %& \multicolumn{2}{c}{Test-2} \\ 
        Strategy & \# Parallels & NLL$\downarrow$ & BLEU$\uparrow$ &
        NLL$\downarrow$ & BLEU$\uparrow$ \\ %& NLL$\downarrow$ & BLEU$\uparrow$ \\
        \hline\hline
        Greedy & 1 & 27.879 & 15.5 & 26.4928 & 16.66 \\ %& 24.0416 & 13.68 \\
        \hline
        NPAD & 5 &   21.5984 & 16.09  &  24.3863 & 17.51  \\ %&  22.1534 & 14.65  \\
        NPAD & 10 &  21.054  & 16.33  &  23.6942 & 17.81  \\ %&  21.6575 & 14.76  \\
        NPAD & \underline{50} & {\bf 20.4463} &  {\bf 16.71}  & \underline{23.0111} &  \underline{18.03} %&   \underline{20.8868} & \underline{14.98} 
    \end{tabular}
\end{minipage}
    \begin{minipage}{0.34\textwidth}
    \caption{Effect of the number of parallel decoding processes. 
        For NPAD, $\sigma_0=0.3$.  }
        \label{tab:n_parallel}
\end{minipage}
\end{table}

\paragraph{Effect of the Number of Parallel Chains}

Next, we see the effect of having more parallel decoding processes of the
proposed NPAD. As we show in Table~\ref{tab:n_parallel}, the translation
quality, in both the average negative log-probability and BLEU, improves as more
parallel decoding processes are used, while it does significantly better than
greedy strategy even with five chains. We observed this trend for all the other
noise levels. This is an important observation, as it
implies that the performance of decoding can easily be improved without
sacrificing the delay between receiving the input and returning the result by
simply adding in more cores/machines. 

\begin{table}[h]
    \small 
    \centering
    \begin{minipage}{0.8\textwidth}
    \begin{tabular}{c c c c || c c | c c}% | c c}
        & Beam       & & \#    & \multicolumn{2}{c|}{Valid} & \multicolumn{2}{c}{Test-1} \\ %& \multicolumn{2}{c}{Test-2} \\ 
        Strategy & Width & $\sigma_0$  & Chains & NLL$\downarrow$ & BLEU$\uparrow$ &
        NLL$\downarrow$ & BLEU$\uparrow$ \\ %& NLL$\downarrow$ & BLEU$\uparrow$ \\
        \hline\hline
        Greedy & 1 & - & 1 & 27.879 & 15.5 & 26.4928 & 16.66 \\ %& 24.0416 & 13.68 \\
        NPAD+G & 1 & 0.3 & 50 & 20.4463 &  16.71  & 23.0111 &  18.03 \\ %&   \underline{20.8868} & \underline{14.98} 
        \hline
        Beam  & 5 & - & 1 &                                  20.1842 &17.03   & 22.8106 &18.56 \\ %&  20.661  & 15.96  \\
        NPAD+B & 5 & 0.3 &5 &        19.8106 & {\bf 17.19}   & 22.1374 &\underline{18.64} \\ %&  20.2932 & 15.86   \\
        NPAD+B & 5 & 0.1 &10 &       {\bf 19.7771} &17.16   & \underline{22.1594} &18.61 \\ %&  20.182  & 16.04  \\
        \hline                                                         
        Beam & 10 & - & 1 &                                 19.9173 &17.13   & 22.4392 &18.59 \\ %&  20.4218 & 16.02    \\
        NPAD+B &  10 & 0.2 &5 &       19.7888 & {\bf 17.16}   & 22.1178 & \underline{18.68} \\ %&  20.1668 & 15.8    \\
        NPAD+B &  10 & 0.1 &10 &      {\bf 19.6674} & 17.14   & \underline{21.9786} &18.78 \\ %&  20.1407 & 15.87   \\
    \end{tabular}
\end{minipage}
    \begin{minipage}{0.19\textwidth}
        \caption{NPAD with beam search (NPAD+B). We report the NPAD+B's with the
        best average log-probability on the validation set.}
        \label{tab:npad_p}
\end{minipage}
\end{table}

\paragraph{NPAD with Beam Search}
As described earlier, NPAD can be used with any other deterministic decoding
strategy. Hence, we test it together with the beam search strategy. As in
Table~\ref{tab:npad_p}, we observe again that the proposed NPAD improves the
deterministic strategy. However, as the beam search is already able to find a
good solution, the improvement is much smaller than that against the greedy
strategy. 

In Table~\ref{tab:npad_p}, we observe that difference between the greedy and
beam search strategies is much smaller when the NPAD is used as an outer loop.
For instance, comparing the greedy decoding and beam search with with 10, the
differences without and with NPAD are 7.9617 vs. 0.7789 (NLL) and 1.66 vs. 0.43
(BLEU). This again confirms that the proposed NPAD has a potential to make the
neural machine translation more suitable for deploying in the real world.

\begin{table}[ht]
\vspace{-2mm}

    \small 
    \centering
    \begin{minipage}{0.8\textwidth}
    \begin{tabular}{c c c c || c c | c c}% | c c}
        & Beam       &  & \#    & \multicolumn{2}{c|}{Valid} & \multicolumn{2}{c}{Test-1} \\ %& \multicolumn{2}{c}{Test-2} \\ 
        Strategy & Width & $\star$   & Chains & NLL$\downarrow$ & BLEU$\uparrow$ &
        NLL$\downarrow$ & BLEU$\uparrow$ \\ %& NLL$\downarrow$ & BLEU$\uparrow$ \\
        \hline\hline
        Beam  & 5 & - & 1 &                                  20.1842 &17.03   & 22.8106 &18.56 \\ %&  20.661  & 15.96  \\
        NPAD+B & 5 & 0.3 &5 &        19.8106 & {\bf 17.19}   & 22.1374 &\underline{18.64} \\ %&  20.2932 & 15.86   \\
        Diverse & 5 & 0.001 & 1 & 20.1859 & 17.03  & 22.8156 & 18.56 \\ %15.97   20.6634 \\
        \hline                                                         
        Beam & 10 & - & 1 &                                 19.9173 &17.13   & 22.4392 &18.59 \\ %&  20.4218 & 16.02    \\
        NPAD+B &  10 & 0.2 &5 &       19.7888 & {\bf 17.16}   & 22.1178 & \underline{18.68} \\ %&  20.1668 & 15.8    \\
        Diverse & 10 & 0.1 & 1 & 19.8908 & 17.2 & 22.4451 & 18.62  \\ % 16.04   20.3892 \\
    \end{tabular}
\end{minipage}
    \begin{minipage}{0.19\textwidth}
        \caption{NPAD vs. diverse decoding. The hyperparameter $\eta_0$ was
            selected based on the BLEU on the validation set. ($\star$) $\sigma_0$ if
    NPAD, and $\eta$ if Diverse.}
        \label{tab:npad_vs_diverse}
\end{minipage}

\vspace{-4mm}
\end{table}

\paragraph{NPAD vs Diverse Decoding}

In Table~\ref{tab:npad_vs_diverse}, we present the result using the diverse
decoding. The diverse decoding was proposed in \cite{li2016mutual} as a way to
improve the translation quality, and accordingly, we present the best approaches
based on the validation BLEU. Unlike what was reported in \cite{li2016mutual},
we were not able to see any substantial improvement by the diverse decoding.
This may be due to the fact that Li~\&~Jurafsky~\cite{li2016mutual} used
additional translation/language models to re-rank the hypotheses collected by
diverse decoding. As those additional models are trained and selected for a
specific application of machine translation, we find the proposed NPAD to be
more generally applicable than the diverse decoding is. It is however worthwhile
to note that the diverse decoding may also benefit from having the NPAD as an
outer loop.

\section{Conclusion and Future Work}

In this paper, we have proposed a novel decoding strategy for conditional
recurrent language models. The proposed strategy, called noisy, parallel
approximate decoding (NPAD), exploits the hidden state space of a recurrent
language model by injecting unstructured Gaussian noise at each transition.
Multiple chains of this noisy decoding process are run in parallel without any
communication overhead, which makes the NPAD appealing in practice. 

We empirically evaluated the proposed NPAD against the widely used greedy and
beam search as well as stochastic sampling and diverse decoding strategies. The
empirical evaluation has confirmed that the NPAD indeed improves decoding, and
this improvement is especially apparent when the inner decoding strategy, which
can be any of the existing strategies, is more approximate. Using NPAD as an
outer loop significantly closed the gap between fast, but more approximate
greedy search and slow, but more accurate beam search, increasing the potential
for deploying conditional recurrent language models, such as neural machine
translation, in practice.

\paragraph{Future Work}

We consider this work as a first step toward developing a better decoding
strategy for recurrent language models. The success of this simple NPAD
suggests a number of future research directions. First, thorough investigation
into injecting noise during training should be done, not only in terms of
learning and optimization (see, e.g., \cite{bengio2015scheduled}), but also in
the context of its influence on decoding. It is conceivable that there exists a
noise injection mechanism during training that may fit better with the noise
injection process during decoding (as in the NPAD.) Second, we must study the
relationship between different types and scheduling of noise in the NPAD in
addition to white Gaussian noise with annealed variance investigated in this
paper. Lastly, the NPAD should be validated on the tasks other than neural
machine translation, such as image/video caption generation and speech
recognition (see, e.g., \cite{cho2015describing} and references therein.)

\subsubsection*{Acknowledgments}

KC thanks the support by Facebook, Google (Google Faculty Award 2016) and NVidia
(GPU Center of Excellence 2015-2016).

%\newpage
\small

\bibliography{npg}
\bibliographystyle{abbrv}

\end{document}